\documentclass[10pt,twocolumn,letterpaper]{article}
\usepackage{cvpr}
\usepackage{times}
\usepackage{epsfig}
\usepackage{graphicx}
\usepackage{amsmath}
\usepackage{amssymb,bbm}
\usepackage{subfig}
\usepackage[hang,flushmargin]{footmisc}
\graphicspath{{Figs/}}
\usepackage{enumitem}
\setitemize{noitemsep,topsep=0pt,parsep=0pt,partopsep=0pt}

\usepackage[pagebackref=true,breaklinks=true,letterpaper=true,colorlinks,bookmarks=false]{hyperref}

\cvprfinalcopy 

\def\cvprPaperID{18} 

\ifcvprfinal\fi

\begin{document}
	
	\title{Privacy Preserving Group Membership Verification and Identification}

\author{Marzieh Gheisari, Teddy Furon, Laurent Amsaleg\\
	Univ Rennes, Inria, CNRS, IRISA, France\\
	{\tt\small \{\texttt{marzieh.gheisari-khorasgani, teddy.furon}\}@\texttt{inria.fr}},
	{\small \texttt{laurent.amsaleg}@\texttt{irisa.fr}}
}

\maketitle
\newcommand\vect[1]{\mathbf{#1}} 
\newcommand\func[1]{\mathsf{#1}} 
\def \group {\mathcal{S}}
\def \x {\vect{x}} 
\def \y {\vect{y}} 
\def \z {\vect{z}} 
\def \f {\vect{f}}
\def \X {\vect{X}}
\def \Y {\vect{Y}}
\def \Z {\vect{Z}}
\def \w {\vect{u}} 
\def \m {\vect{m}}
\def \e {\vect{e}}
\def \h {\vect{e}}  
\def \a {\vect{a}}
\def \r {\vect{r}}
\def \n {\vect{n}} 
\def \W {\vect{W}} 
\def \G {\vect{G}} 
\def \real {\mathbb{R}} 
\def \dim {d} 
\def \rep {\vect{r}} 
\def \hyp {\mathcal{H}} 
\def \one {\vect{1}}
\def \W {\vect{W}}
\def \H {\vect{E}}
\def \R {\vect{R}}
\def \Av {\vect{A}}
\def \A {\mathcal{A}}
\def \Pr {\mathbb{P}} 
\def \Pfn {P_{\mathsf{fn}}}
\def \Pfp {P_{\mathsf{fp}}}
\def \Ptp {P_{\mathsf{tp}}}
\def \pfn {p_{\mathsf{fn}}}
\def \ptp {p_{\mathsf{tp}}}
\def \pfp {p_{\mathsf{fp}}}
\def\un{{\mathbbm{1}}}
\def\E{{\mathbbm{E}}}
\def\Var{{\mathbbm{V}}}
\def\0{{\vect{0}}}
\def\etal{{\it et al.}}
\def\ie{{\it i.e.}}
\def\AoE{\text{\tiny{AoE}}}
\def\EoA{\text{\tiny{EoA}}}

\newcommand\alert[1]{{\color{red}{#1}}}

\newcommand\blfootnote[1]{%
	\begingroup
	\renewcommand\thefootnote{}\footnote{#1}%
	\addtocounter{footnote}{-1}%
	\endgroup
}

\maketitle
\thispagestyle{empty}
\begin{abstract}
When convoking privacy, group membership \emph{verification} checks if a biometric trait corresponds to one member of a group without revealing the identity of that member.  Similarly, group membership \emph{identification} states which group the individual belongs to, without knowing his/her identity.
 A recent contribution provides privacy and security for group membership protocols through the joint use of  two mechanisms: quantizing biometric templates into discrete embeddings, and aggregating several templates into one group representation.

 This paper significantly improves that contribution because it jointly learns how to embed and aggregate instead of imposing fixed and hard coded rules. This is demonstrated by exposing the mathematical underpinnings of the learning stage before showing the improvements through an extensive series of experiments targeting face recognition. Overall, experiments show that learning yields an excellent trade-off between security / privacy and the verification / identification performances.

\end{abstract}
\blfootnote{Research supported by the ERA-Net project ID\_IoT 20CH21\_167534.}

\section{Introduction}
\label{sec:Introduction}

The verification that an item, a device or an individual is a member of a group is a natural task which forms the basis of many applications granting or refusing access to sensitive resources (buildings, wifi, payment, conveyor units, \ldots). Group membership can be implemented through a two-phase process where an \emph{identification} is first performed, revealing the identity of the individual under scrutiny, followed by a \emph{verification} phase where it is checked whether or not the identified individual is indeed a member of the claimed group. That implementation breaks privacy: there is no \emph{fundamental} reason to identify the individual before running the verification step. It is easier, but not truly needed. It is fundamental to distinguish the members of the group from the non-members, but it does not require to distinguish members from one another.

The same comments hold for group \emph{identification}.
Such a system manages multiple groups, separating \eg\ individuals according to the team they work in. The goal is then to identify the precise group a member belongs to, without proceeding first to the identification of the individual.

Privacy preserving 
protocols exist, for the most part in the area of signal processing in the encrypted domain~\cite{Troncoso-Pastoriza:2013hi,Erkin:2009cz,Sadeghi:2010bl}. These protocols, however, are costly, making them hard to use in practice.

In contrast, some other approaches propose privacy preserving mechanisms that are cheaper to use because they solely rely on pure signal processing techniques, excluding all encryption mechanisms. They tend to be less secure, of course, but they put in the way of malicious users so many obstacles that breaking privacy or counterfeiting identities becomes discouragingly complicated and almost impossible. At their core, these approaches use for example sketching or quantization techniques that prohibit reconstructing identities with sufficient precision~\cite{Bianchi:2012sy,Boneh:2007om}. Other techniques embed real vectors into another representational space with the purpose of security and privacy~\cite{Razeghi2017wifs, Razeghi2018icassp}.

Recently, Gheisari \etal{} proposed a privacy preserving group membership verification protocol quantizing identities' biometric templates into discrete embeddings and aggregating multiple embeddings into a group representation~\cite{Gheisari2019icassp}. That scheme has several desirable properties: It is pure signal processing and linear algebra, hence it is cheap to run; quantization and aggregation fully succeed to make reconstruction difficult and impede identification; it is demonstrated to allow trading-off the strength of its security against group verification error rates.

That work, however, is fully deterministic in the sense that it sticks to a set of hard coded rules that drive the way templates are embedded, how they are grouped and then aggregated into group representations. Although well justified and sound, these rules govern the behavior of two independent procedures, one for embedding, the other for aggregating.
The main contribution of this paper shows that jointly considering the embedding and aggregation stages results in better performances, \ie\ a better membership verification without damaging security.

This paper is therefore revisiting the core mechanisms proposed by Gheisari \etal{} in~\cite{Gheisari2019icassp}, by \emph{jointly} optimizing the embedding and aggregation stages. It is structured as follows. Section~\ref{sec:RelatedWorks} defines the context of this work and surveys existing approaches. Then, Section~\ref{sec:starting-point} 
gives a formal problem statement and the details of proposed method are introduced in Section~\ref{sec:ProposedMethod}. Section~\ref{sec:Experiments} presents the experimental protocol and results.



\section{Context and Related Work}
\label{sec:RelatedWorks}

\subsection{Context}
Without any loss of generality, group membership verification needs first to
acquire \emph{templates} of items (passive PUF), devices (active PUF),  or individuals (biometric trait) and to enroll them  into a data structure stored in a server. Then, at verification time,  that data structure is queried by a client with a new template and the access is granted or refused. Security assesses that the data structure is adequately protected so that a honest but curious server cannot reconstruct the templates. Privacy requires that verification should proceed without disclosing the identity.

The nature of templates can vary from one application to the other.  For example, the templates encode information related to fingerprints, iris, or faces of individuals, or to Physically Unclonable Functions (PUF) like speckle patterns captured from laser-illuminated transparent plastic.

It is worth highlighting two fundamental properties of the templates.  The template used at verification time is a noisy version of the one acquired at enrollment time.  Lighting conditions, blood pressure, aging, worn-outs, transient physical conditions are possible factors that might cause variations at acquisition time. The verification protocol must be able to absorb such variations and cope with the continuous nature of the templates. However, it is very unlikely that a noisy version of the template corresponding to one group member gets similar enough to the enrolled template of any other group member. The first property is therefore in relation with the continuous and distinguishable nature of the templates.  The second property is in relation with the statistical independence of the enrolled templates.

The traditional operational scenario considers a server which runs the group membership verification. The server receives queries from clients. A client acquires a new template and then queries the server. Clients are trusted. The server is honest but curious: It may try to reconstruct the enrolled templates or spy on the queries. The design intends to prevent the server from reconstructing the private template from the system while correctly determining whether or not a user is a member of the claimed group (group verification) or identifying the group of membership (group identification).

\subsection{Related Work}
\subsubsection{Cryptography}
\label{sec:Crypto}
There exist cryptographic protocols setting a key management system to provide the members of a group with anonymous authentication~\cite{Schechter:1999qy}.  There is no enrollment of biometric or PUF templates since membership is equivalent to holding one of the valid keys.  As for biometry or PUF applications, our scenario is different from authentication, identification and secret binding.  These applications secure the templates at the server and/or the client sides, but ultimately reveal the identity of the user/object.

Signal processing in the encrypted domain can provide a solution to group membership verification.  At enrollment time, each template is quantized and protected with homomorphic encryption.  The query is protected in the same manner at the verification stage.  This allows to compute distances between the query and the templates in the encrypted domain~\cite{Troncoso-Pastoriza:2013hi}.  There exist protocols for comparing encrypted results to a threshold~\cite{Erkin:2009cz,Sadeghi:2010bl}.  These encrypted comparisons are sent back to the clients which decrypt and check whether there is at least one positive.  Security and privacy are as high as the security of the cryptographic primitives.  Homomorphic encryption, however, is costly, both in terms of memory footprint (space) and CPU consumption (time).

\subsubsection{Sketching techniques}
Group membership is linked to the well-known Bloom filter that is used to test whether an element is a member of a set.  A Bloom filter hashes and blends elements into one array of bits.
When used in the context of privacy, it was demonstrated that a server using Bloom filters cannot infer any information on one specific entry~\cite{Bianchi:2012sy}.
However, a Bloom filter can not be used as is in our application, but it needs two adaptations.  First, a Bloom Filter deals with discrete objects whereas we consider continuous high dimensional vectors, the templates.  We first need to design a quantizer hashing the continuous templates into discrete objects.  That quantizer must absorb the noise, \ie{} the difference between the enrolled and the fresh template.  Second, at verification time, the hash of the query cannot be sent in the clear for privacy reasons~\cite{Boneh:2007om}.  For instance, Beck and Kerschbaum protect the query with partially homomorphic encryption since there is no need to protect the filter at the server side~\cite{Beck:2013df}.

\subsubsection{Aggregating, Embedding}

Aggregating signals into one representation is a very common mechanism in computer vision.  Approaches like BoW (Bag of Words)~\cite{Sivic:2003qp}, VLAD~\cite{jegou:inria-00633013}, Fisher vectors~\cite{Perronnin:2007qm}, aggregate some local descriptors extracted from one image into one global description.  However, these approaches are designed to facilitate the identification of similar elements in images and have no security or privacy capabilities.

Another recent aggregation method better fits with the security and privacy requirements that we need. In~\cite{iscen:hal-01481220}, Iscen \etal{} use the \emph{group testing} paradigm to design a strategy for packing a random set of image descriptors into a unique high-dimensional vector. One salient property of that strategy is that the similarity between images can be determined by solely comparing these (few) aggregated vectors to the description of the query, without the need of the original (and numerous) raw image descriptors. This saves space (memory footprint of the database) and time (complexity at query time). These gains are the main motivation of~\cite{iscen:hal-01481220}.

As for the embedding, our scenario is similar to a privacy-preserving identification mechanism based on sparsifying transform~\cite{Razeghi2017wifs, Razeghi2018icassp}. It obtains a sparse ternary embedding preserving locality information while ensuring privacy of the data users and security of the templates.

Indeed, paper~\cite{Gheisari2019icassp} uses the aggregation methods designed in~\cite{iscen:hal-01481220} combined with the sparsifying transform~\cite{Razeghi2017wifs, Razeghi2018icassp}.  Our design resorts to this latter mechanism for inheriting its privacy-preserving properties as well, yet we replace the former ad-hoc hard coded rules of~\cite{iscen:hal-01481220} by a machine learning approach.

\subsubsection{Face recognition}
Aggregating templates into one group representation also exists in face recognition, but privacy and security are not a requirement in the design of this primitive. They are usually taken into account with an extra layer (see Sect.~\ref{sec:Crypto}). 
In a common approach, multiple face captures of the same person  are combined to gain robustness against poses, expression, and quality variations~\cite{Zhong:2018ab}. In our scenario, the group is composed of unique faces of different persons.
  
Paper~\cite{Zhong:2018aa} computes a single compact descriptor for the faces of celebrities appearing in the same picture. The query is a small set of face descriptors and their system returns photos where these celebrities appear altogether.  This paper consists in jointly learning the face description and the aggregation mechanism.  The authors report ``\textit{a minimal loss of discriminability up to two faces per image, and degrades slowly after that}''.

Our paper deals with the aggregation for a given on-the-shelf face descriptor. On one hand, our setup is easier because the query is a single face. On the other hand, our group typically comprises more than two faces, and each is possibly captured according to very different conditions.



\section{Problem Formulation}
\label{sec:starting-point}
This section details the way the group membership problem is formulated and also defines the metrics used to evaluate performances.

\subsection{Notations}

The set of individuals is denoted as $[N]:=\{1,\ldots,N\}$, partitionned into $M$ groups: $[N] = \bigcup_{g\in[M]}\group_{g}$.  The size of a group $\group$ is denoted by $|\group|$.
In the experiments of Sect.~\ref{sec:Experiments}, all the groups share the same size denoted as $m$.

The templates to be enrolled are the vectors $\{\x_{1}$, \ldots, $\x_{N}\}\subset \real^{\dim}$.  They are stored column wise in $d\times N$ matrix $\X$ according to the partition over the groups: $\X=[\X_{1},\ldots, \X_{M}]$, where $\X_{g}$ stores the templates belonging to $\group_{g}$, $\forall g\in[M]$. The output of the enrollment is a $\ell\times M$ matrix $\R = [\r_{1},\ldots, \r_{M}]$ composed of the representations of the $M$ groups.  It is imposed that the group representations are quantized and sparse: $\r_{g}\in\A^{\ell}$ with $\A:=\{-1,0,1\}$ and $\|\r_{g}\|_{0}\leq S<\ell$, $\forall g\in[M]$.  Moreover, the analysis is restricted to the case where $\ell \leq d$.

The template for which the membership has to be verified is a query vector $\y\in\real^{\dim}$ that is cast onto $\A^{\ell}$ thanks to an embedding before being compared to the group representations. 
 
\subsection{Embedding}
\label{sub:Embedding}
Function $\func{e}:\real^{\dim}\to \mathcal{A}^{\ell}$ maps a vector to a sequence of $\ell$ discrete symbols.  We intentionally choose the sparsifying transform coding described in~\cite{Razeghi2017wifs, Razeghi2018icassp} for its security and privacy good properties.  It projects $\mathbf{x} \in \real^{d}$ on the column vectors of $\mathbf{W} \in \real^{\dim \times \ell}$.  The output alphabet $\mathcal{A}=\{-1, 0, +1 \}$ is imposed by quantizing the components of $\W^{\top}\mathbf{x}$.  The $\ell-S$ components having the lowest amplitude are set to 0.  The $S$ remaining ones are quantized to +1 or -1 according to their sign.
\begin{eqnarray}
\label{eq:EmbedEncoding}
\func{e}: \real^{\dim}&\to& \A^{\ell}\\
\x &\mapsto& \func{e}(\x) = \func{T}_{S}(\W^T\x).\nonumber
\end{eqnarray}

\subsection{Performances Metrics}
Paper~\cite{Gheisari2019icassp} defines metrics to assess (i) the ability of the protocol to correctly perform the verification task, and (ii) the security and privacy.

\subsubsection{Verification Performance}

The first metrics is seeded by considering the following two hypotheses:
\begin{itemize}
\item $\hyp_{1}$: The query is related to one of the vectors of group $g$. For instance, it is a noisy version of vector $j$, $\y=\x_{j}+\n$, with $\n$ to be a noise vector and $j\in\group_g$.
\item $\hyp_{0}$: The query is not related to any vector in the group.
\end{itemize}

The group membership test decides which hypothesis is deemed true by comparing $\func{e}(\y)$ to~$\rep_g$.
This is done by first computing a score function $\func{c}$ and thresholding its result: $t:=[\func{c}(\func{e}(\y),\rep_{g})>\tau]$.


The probabilities of false negative, $\pfn(\tau):=\Pr(t=0|\hyp_{1})$, and false positive, $\pfp(\tau):= \Pr(t=1|\hyp_{0})$ are summarized
by the AUC (Area Under the ROC Curve). Paper~\cite{Gheisari2019icassp}  also considers $\pfn(\tau)$ for $\tau$ s.t. $\pfp(\tau)=\epsilon$,  a required false positive level, $0.05$ for example.

\subsubsection{Security and Privacy Performance}
A curious server can reconstruct the query template $\y$ from its embedding: $\hat{\y} := \func{rec}(\func{e}(\y))$.
This endangers privacy of the querying user.
The mean squared error assesses how accurate is this reconstruction:
\begin{equation}
\label{eq:MSE_P}
\func{MSE}_P = d^{-1}\E(\|\Y -\func{rec}(\func{e}(\Y))\|^{2}),
\end{equation}
assuming that $\Y$ is a white gaussian vector in $\real^{\dim}$.

As for the security of the enrolled templates, a curious server can only reconstruct a single vector $\hat{\x}$ from the aggregated representation, and this vector serves as an estimation of any template in the group:
\begin{equation}
\label{eq:MSE_S}
\func{MSE}_{S} = (dN)^{-1}\sum_{g=1}^{M}\sum_{i=1}^{|\group_{g}|}\E(\|\x_{i}  -\hat{\x}_{g}\|^{2}).
\end{equation}


\section{Proposed Method} 
\label{sec:ProposedMethod}

\subsection{Variants of the Protocol} 
Our group membership protocol is based on embedding and aggregation functions. Whereas the embedding function $\func{e}$ is fixed, we can have two constructions:
The aggregation of embeddings or the embedding of the aggregation. 




\paragraph{EoA} aggregates raw templates into one vector of $\real^\dim$, and then embeds this vector.
Then, the group representative vector $\rep_{g}$ is computed as:
\begin{eqnarray}
\rep_{g}&=&\func{e}(\func{a}_{\EoA}(\X_{g})),
\end{eqnarray}
where $\X_{g}$ is the $\dim\times|\group_{g}|$ matrix storing the templates of the $g$-th group.

\paragraph{AoE} first embeds each template according to (\ref{eq:EmbedEncoding}) before aggregating them into $\rep_{g}$:
\begin{eqnarray}
\rep_{g}&=& \func{a}_{\AoE}\left(\{\func{e}(\x_{i})\}_{i\in\group_{g}}\right).
\end{eqnarray}



This work aims at learning the aggregated vectors and the embeddings \emph{jointly}.
For both construction, this is done by minimizing an objective function summing a cost for embedding $C^{E}$ and a cost for aggregating $C^{A}$.

For AoE (first embed, then aggregate), denote $\H\in\A^{\ell\times N}$ the matrix storing the embeddings of the enrolled templates.  Like for $\X$, we write $\H:= [\H_{1}, \ldots, \H_{M}]$ with $\H_{g}$ the matrix gathering the embeddings of the templates of group $\group_{g}$.  For EoA (first aggregate, then embed), denote $\Av := [\a_{1}, \ldots, \a_{M}]\in\real^{\dim\times M}$ the matrix gathering the aggregations of the templates enrolled in a group.

Matrices $\H$ and $\Av$ will be defined through optimization problems detailed below.  For the embedding, function $\func{e}$ is still prototyped according to Sect.~\ref{sub:Embedding}.
Papers~\cite{Razeghi2017wifs, Razeghi2018icassp} show that privacy and security stem from the sparsifying transform.  Only its matrix $\W$ is learned.

\begin{figure}
	\centering
	\def\svgwidth{0.8\columnwidth}
	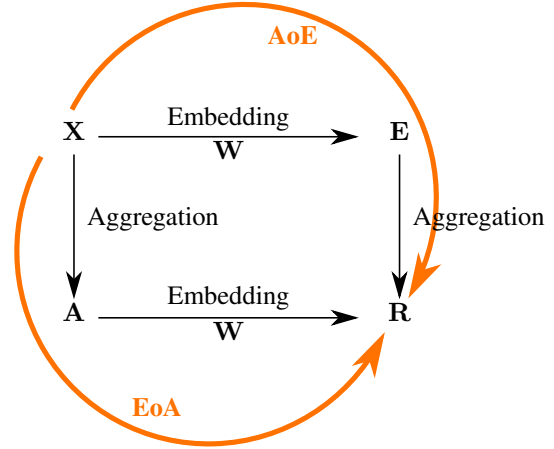
	\caption{Overview}
	\label{fig:Overview}
	\vspace{-10pt}
\end{figure}

For EoA, Fig.~\ref{fig:Overview} shows that it starts from $\X$ to create $\Av$ before outputting $\R$ using matrix $\W$. This defines the optimization problem:

\begin{equation}
\min_{\Av,\W,\R} \gamma C_{\EoA}^{A}(\X,\Av) + C_{\EoA}^{E}(\Av,\R,\W),
\label{eq:OptimizationProblemEoA}
\end{equation}
where $\gamma$ is the penalty parameter, $\Av\in\real^{\dim\times M}$, $\W\in\real^{\dim\times \ell}$, and $\R\in\A^{\ell\times M}$.

On the other hand, Fig.~\ref{fig:Overview} shows that for AoE, we start from $\X$ to create $\H$ using matrix $\W$ before outputting the group representations $\R$. This defines the optimization problem:
\begin{equation}
\min_{\H,\W,\R} C_{\AoE}^{E}(\X,\W,\H) + \xi C_{\AoE}^{A}(\H,\R),
\label{eq:OptimizationProblemAoE}
\end{equation}
with $\H\in\A^{\ell\times M}$, $\W\in\real^{\dim\times \ell}$, and $\R\in\A^{\ell\times M}$.

Under both constructions, 
the optimization is joint because the embedding and the aggregating costs share a common variable ($\Av$ or $\H$).  What follows define the costs and solve the optimization problems.


\subsection{AoE: Aggregation of Embeddings}
This scheme first embeds and then aggregates by solving~\eqref{eq:OptimizationProblemAoE}.
The cost for embedding is defined as
\begin{eqnarray}
\label{eq:QuantizationLoss}
C_{\AoE}^{E}(\X,\W,\H)& :=&  \sum_{i=1}^{N} \left\Vert\h_i -\W^\top \x_i \right\Vert^2_2,\\
&=& \|\H - \W^{\top}\X\|_{F}^{2}.
\end{eqnarray}
This term represents the fidelity penalty of an embedding $\h$ w.r.t. a template $\x$ in the transformed domain.
 
For each group of embedded templates, the aggregated vector should satisfy some properties as well:
\begin{itemize}
	\item For each group the overall distance between group members and the aggregated vector is minimized. 
	\item Aggregated vector should be represented as a sparse ternary code.
\end{itemize}
The cost of aggregation is then defined as
\begin{equation}
  \label{eq:m}
C_{\AoE}^{A}(\H,\R) := \sum_{g=1}^{M} \|\H_{g} - \rep_{g}\one_{|\group_g|}^{\top}\|^2.
\end{equation}
We add the following constraints:%
\begin{eqnarray}
\W^\top\W&=&\vect{I}_\ell,\\
\left\Vert \e_i \right\Vert_0&\leq& S,\; \forall  i \in [N]\\
\left\Vert \rep_g \right\Vert_0&\leq& S \; \forall g \in [M].
\end{eqnarray}
The constraint makes sure that the representative $\rep$ is sparse, ternary, and diverse.

We propose to optimize problem~\eqref{eq:OptimizationProblemAoE} iteratively by alternating updates of one parameter while fixing the remaining ones.
Each step minimizes the total cost function lower bounded by 0, insuring a convergence to a local minimum.

\paragraph{$\W$-Step.} We fix $\H$ and $\R$ and update $\W$ by solving:
\begin{equation}
\label{eq:OptW}
\begin{aligned}
& \underset{\W} {\min}
& &  \left\Vert\H -\W^\top \X \right\Vert^2_F\\
&\text{s.t.}
& & \W^\top\W=\vect{I}_\ell
\end{aligned}
\end{equation}
This problem is a least square Procruste problem with orthogonality constraint.
By setting $\vect{S}:=\X\H^\top$,~\cite{Schonemann1966} shows that $\W=\vect{UV}^\top$, where $\vect{U}$ contains the eigenvectors corresponding to the $\ell$  ($\ell < \dim$) largest eigenvalues of $\vect{SS}^\top$ and $\vect{V}$ contains the eigenvectors of $\vect{S}^\top\vect{S}$.
\paragraph{$\H$-Step.} $\W$ and $\R$ being fixed, we can solve the problem for each $\H_{g}$ independently: $\forall g\in[M]$,
\begin{equation}
\label{eq:OptHp}
\begin{aligned}
& \underset{\H_g} {\min}
& & \left\Vert\H_g -\W^\top \X_g \right\Vert^2_F
+\xi \left\Vert  \H_g -\rep_g \one_{|\group_g|}^\top\right\Vert^2_F\\
& \text{s.t.}
& & \H_g \in \A^{\ell \times |\group_{g}|} ,\; \left\Vert \h_i \right\Vert_0\leq S, \; \forall i \in \group_{g}.
\end{aligned}
\end{equation}
According to~\cite{Razeghi2017wifs}, we first find the solution without considering the constraints and then apply ternarization function $\func{T}_S$~\eqref{eq:EmbedEncoding} to obtain sparse codes. Therefore $\H_g$ is found as:
\begin{equation}
\label{eq:OptHs}
\begin{aligned}
& & \H_g=\func{T}_S(\W^\top\X_g+\xi \rep_g\one_{|\group_g|}^\top).
\end{aligned}
\end{equation}

\paragraph{$\R$-Step.}
Like for the $\H$-step, updating each group representation $\rep_g$ is done independently, while fixing $\W$ and $\H$:
\begin{equation}
\label{eq:OptRs}
\begin{aligned}
& \underset{\rep_g} {\min}
& & \left\Vert  \H_g -\rep_g \one_{|\group_g|}^\top\right\Vert^2_F\\
& \text{s.t.}
& & \rep_g \in \A^{\ell}, \; \; \left\Vert \rep_g \right\Vert_0\leq S. 
\end{aligned}
\end{equation}
Then the representative of $g$-th group is obtained as:
\begin{equation}
\label{eq:OptHs}
\begin{aligned}
& & \rep_g=\func{T}_S(\H_g\one_{|\group_g|}).
\end{aligned}
\end{equation}

\subsection{EoA: Embedding of Aggregation}
We now consider the construction of~\eqref{eq:OptimizationProblemEoA} that first aggregates and then embeds. 
The cost of the aggregation is defined as:
\begin{equation}
\label{eq:EoA}
\begin{aligned}
C_{\EoA}^{A}(\X,\Av)  := \sum_{g=1}^{M}\left\Vert  \X_g^\top\a_{g} - \one_{|\group_g|}  \right\Vert^2_2 + \eta  \left\Vert \a_{g} \right\Vert^2_2.
\end{aligned}
\end{equation}
Minimizing this cost amounts to equalize the similarity between each members of the group and the aggregated vector $\a$.
The cost for embedding is defined as previously:
\begin{equation}
C_{\EoA}^{E}(\Av,\R,\W) := \|\R - \W^{\top}\Av\|_{F}^{2}.
\end{equation}
As for the constraints:
\begin{eqnarray}
\W^\top\W&=&\vect{I}_\ell, \\
 \rep_{g}\in\A^{\ell}, \left\Vert\rep_g \right\Vert_0&\leq& S, \; \forall g \in [M].
\end{eqnarray}


The optimization problem~\eqref{eq:OptimizationProblemEoA}
with these costs and constraints is solved by iterating the following steps.

\paragraph{$\W$- Step.}
Like for \eqref{eq:OptW}, updating $\W$ while $\vect{R}$, $\vect{A}$ are fixed is a Procruste problem under orthogonality constraint:
\begin{equation}
\label{eq:OptW2}
\begin{aligned}
& \underset{\W} {\text{min}}
& &  \left\Vert\R -\W^\top \Av  \right\Vert^2_F,\\
&\text{s.t.}
& & \W^\top\W=\vect{I}_\ell.
\end{aligned}
\end{equation}
Similar to~\eqref{eq:OptW}, we define $\vect{S}:=\vect{AR}^\top$.
The solution is found as $\W = \vect{UV}^\top$, where $\vect{U}$ contains the eigenvectors corresponding to the $\ell$ largest eigenvalues of $\vect{SS}^\top$ and $\vect{V}$ the eigenvectors of $\vect{S}^\top\vect{S}$.

\paragraph{$\Av$- Step.}
When fixing $\W$ and $\vect{R}$, the aggregated vector for each group $g\in[M]$ is found independently by minimizing:
\begin{equation}
\label{eq:OptAp2}
\underset{\vect{a}_g} {\min}
\left\Vert\rep_g - \W^\top\a_g \right\Vert^2_2
+ \gamma (\left\Vert  \X_g^\top\a_g - \one_{|\group_g|}  \right\Vert^2_2 + \eta  \left\Vert \a_g \right\Vert^2_2),
\end{equation}
whose solution is
\begin{equation*}
\label{eq:OptAs2}
\a_g={(\W\W^\top+\gamma(\X_g\X_g^\top+\eta \vect{I}_\dim))}^{-1} 
(\W\rep_g+\gamma \X_g \one_{|\group_g|}).
\end{equation*}
\paragraph{$\R$- Step.}
Projection matrix $\W$ and the group aggregations $\Av$ are fixed.
The group representatives are obtained by applying sparse ternarization function on the projected aggregated vectors:  $\R = \func{T}_S(\W^\top \Av)$.

\begin{figure*}[tb]%
	\centering
	\includegraphics[width=0.95\linewidth]{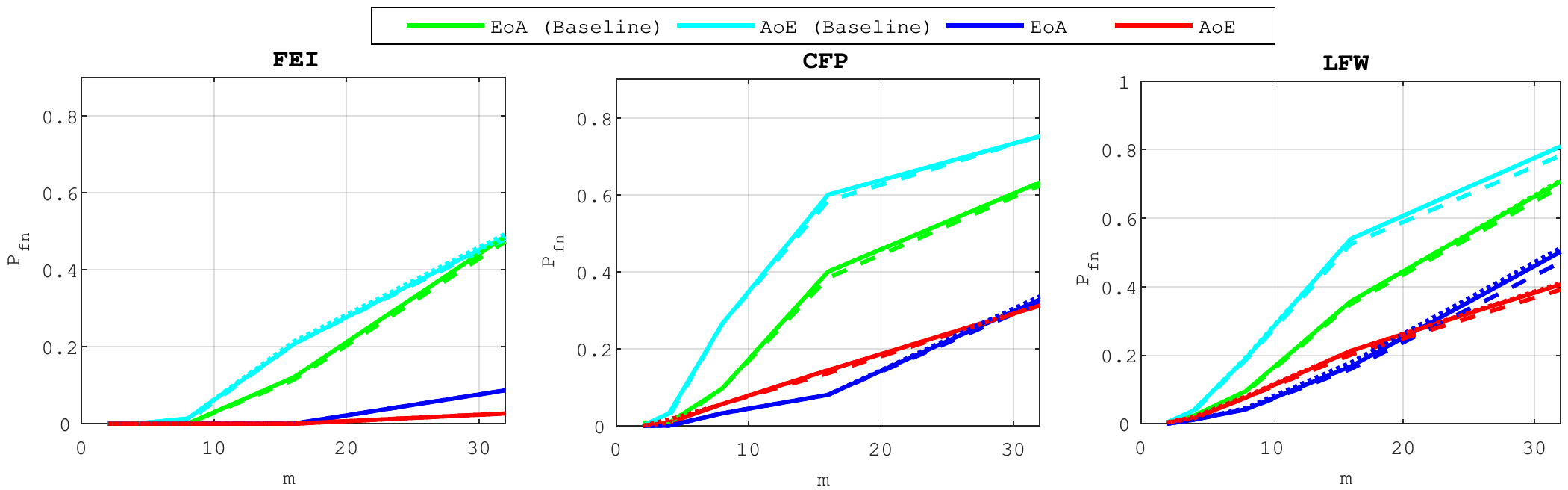}
	\caption{Performances comparison of our schemes \textit{vs.} baseline for varying group size $m$. $\Pfn$ at $\Pfp=0.05$ for group verification (solid), the first step of group identification (dashed), and $P_{\epsilon}$ for the second step of group identification (dotted).} 
	\label{fig:baseline}%
      \end{figure*}

\section{Experiments}
\label{sec:Experiments}
We fully implemented the group membership protocol that is described in~\cite{Gheisari2019icassp}.
Experimenting with this implementation gives the baseline performances.
Note that the experimental part of~\cite{Gheisari2019icassp} only deals with synthetic data.
Our implementation could reproduce these results. However, the sequel presents comparisons on real data.

\subsection{Experimental Setup}
We evaluate the performances of our scheme with face recognition.  Face images are coming from LFW~\cite{huang2008labeled}, CFP~\cite{sengupta2016frontal} and FEI~\cite{thomaz2010new} databases. 
Face descriptors are obtained from a pre-trained network based on VGG-Face architecture~\cite{parkhi2015deep}.  The output vector of the penultimate layer (\ie\ before the final classifier layer) is PCA reduced to a lower dimension ($\dim = 1,024$ for LFW and CFP, $\dim = 256$ for FEI database), and then $L_{2}$-normalized. The result is the template $\x\in\real^{\dim}$.
The values of $\ell$, $S$, $\xi$, $\gamma$ and $\eta$ are set empirically as $0.9d$, $0.7l$, $1$, $10^4$ and $1$ respectively.
Also, not all individuals from these databases are enrolled.
\paragraph{LFW.}
Labeled Faces in the Wild contains 13,233 images of faces collected from the web. We used cropped LFW images.
The enrollment set consists of $N = 1,680$ individuals with at least two images in the LFW database. One random template of each individual is enrolled in the system, playing the role of $\x_{i}$.
The other templates are used for queries. These are partitioned into two subsets: templates that are correlated with $\x_{i}$ with a similarity bigger than $0.95$ form the ``easy queries'' set ; the remaining templates with a similarity bigger than $0.9$ form the ``hard queries'' set.
The remaining individuals not enrolled in the system ($5,749 - N$) play the role of impostors (hypothesis $\hyp_0$).

\paragraph{CFP.}
The Celebrities in Frontal-Profile (CFP) database is composed of $500$ subjects with $10$ frontal and $4$ profile images for each subject in a wild setting. We only use the frontal images.
The impostor set is a random selection of $100$ individuals.
One random template of the remaining individuals is enrolled in the system.
Like the setting described for LFW, we have two subsets of queries.
\paragraph{FEI.}
We use frontal and pre-aligned images of the Brazilian FEI database. There are $200$ subjects with two frontal images (one with a neutral expression and the other with a smiling facial expression).
The database is created by random sampling $150$ individuals. For each identity, one random image is enrolled while the other is used as query. The remaining individuals are considered as impostors.

At the enrollment phase, all groups have exactly the same number of members: $|\group_g| = m, \forall g\in[M]$.
Individuals are randomly assigned to a group.



\begin{figure*}[tb]%
	\centering
	\includegraphics[width=0.95\linewidth,height=5cm]{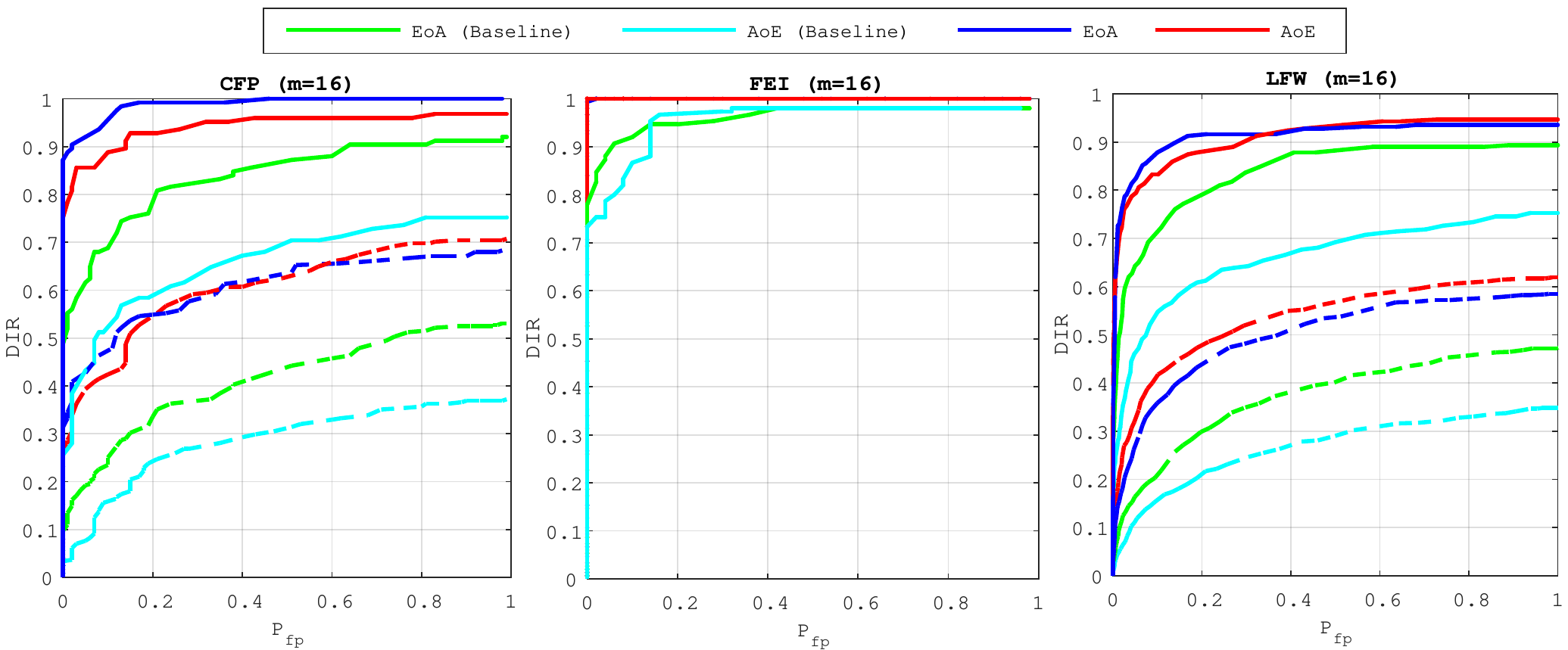}
	\caption{The Detection and Identification Rate ($DIR$) \vs\ $\Pfp$ for group identification.
	Performances for hard queries are plotted in dashed lines.}
	\label{fig:DIR}%
\end{figure*}

The performances of the system are gauged with error probabilities evaluated by Monte Carlo estimator over the testing set.  Since $N$ is not so large, the confidence interval at $95\%$ is $\approx \pm N^{-1/2}=\pm 2.5\%$, which prevents us from estimating small probabilities.  Therefore, \emph{we put our system under stress by selecting a hard setup}.
First, note that LFW and CFP are difficult datasets due to the `in the wild' variations (poses, illuminations, expressions and occlusions). They do not reflect the application of accessing a building (as mentioned in the Introduction) where the capture environment is more under control and the individuals collaborate.  Second, not only the dimension of the templates have been reduced but also the length of the embeddings and the group representation ($\ell = 0.9d$) with a sparsity of $S = 0.7l$ (unless stated otherwise).  Probabilities of errors are then big but measurable with accuracy.  We believe that this protocol makes sense to benchmark approaches.

Two applications scenarios are investigated: group verification and group identification.

\paragraph{Group verification.}
A user claims she/he belongs to group $g$.  This claim is true under hypothesis $\hyp_{1}$ and false under hypothesis $\hyp_{0}$ (\ie\ the user is an impostor).  Her/his template $\y$ is embedded, and $(\func{h}(\y), g)$ is sent to the system, which compares $\func{h}(\y)$ to the group representation $\r_{g}$.  The system accepts ($t = 1$) or rejects ($t=0$) the claim. This is a two hypothesis test with two probabilities of errors: $\Pfp:=\Pr(t=1|\hyp_{0})$ is the false positive rate and $\Pfn:=\Pr(t=0|\hyp_{1})$ is the false negative rate.  The figure of merit is $\Pfn$ when $\Pfp = 0.05$.

\paragraph{Group identification.}        
The scenario is an open set identification where the querying user is either enrolled or an impostor.  The system has two steps. First, it decides whether or not this user is enrolled.  This is  verification as above, except that the group is unknow: The system computes $\delta_{j} = \|\func{h}(\y)-\rep_{j}\|$, $\forall j\in[M]$.  The system accepts $(t=1)$ if the minimum of these $M$ distances is below a given threshold $\tau$.  The figure of merit is $\Pfn$ when $\Pfp = 0.05$.

When $t=1$, the system proceeds to the second step. The estimated group is given by $\hat{g} = \arg \min_{j\in[M]} \delta_{j}$. The figure of merit for this second step is $P_{\epsilon}:= \Pr(\hat{g}\neq g)$ or the Detection and Identification Rate $DIR := (1-P_{\epsilon})(1-\Pfn)$.

\subsection{Exp. \#1: Comparison to the baseline}

Figure~\ref{fig:baseline} shows that our method brings improvement compared to the baseline, since the AoE and the EoA plots are way below the ones corresponding to the baseline. The high probabilities of false negatives for the baseline are caused by the great losses in information: AoE (Baseline) looses information from each template it embeds before the aggregation---the accumulated losses are therefore great; EoA (Baseline) has better performances since plain templates are first aggregated before running the embedding step which causes less information loss.

Our method does not suffer that much from this information loss: EoA and AoE have roughly similar performances, with much more acceptable $\Pfn$ values.

\subsection{Exp. \#2: Detection and Identification Rate}
  
\begin{figure}[tb]%
	\centering
	\includegraphics[width=0.95\linewidth]{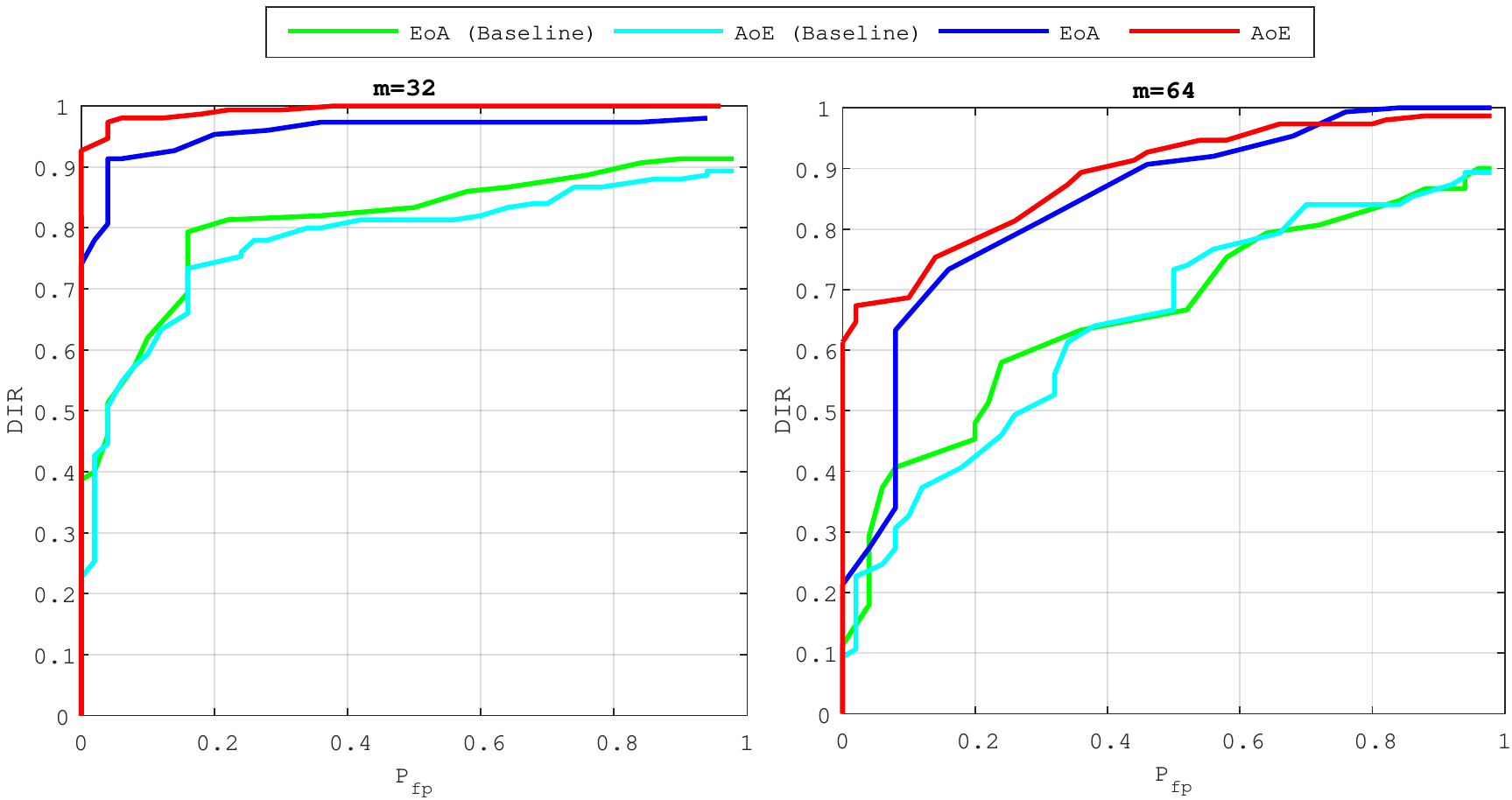}
	\caption{The Detection and Identification Rate ($DIR$) \vs\ $\Pfp$ for group identification on FEI.}
	\label{fig:DIR2}%
\end{figure}
    
Figure~\ref{fig:DIR} compares the $DIR$ performances for group identification with $m=16$.
Our schemes have results close to perfection on the FEI dataset.
Easy queries are correctly handled on CFP but not on the LFW dataset at this size of group.
Hard queries are more difficult to cope with. This is explained by the poor correlations they have with their corresponding  $\x_{i}$. That poor correlation, already existing on the original templates, before any embedding or aggregation, can only lower the performances of any membership identification scheme.

Figure~\ref{fig:DIR2} shows the impact of the size of group on $DIR$.
Packing more templates into one group representation is detrimental even if the queries are well correlated with their corresponding enrolled template.
This suggests to split large groups into subgroups of size lower or equal to $m=32$.
This restricts privacy to $m$-anonymity as the server is now able to identify the subgroup a query belongs to.


\subsection{Exp. \#3: Easy vs. Hard Queries}
\begin{figure}[tb]%
	\centering
	\includegraphics[width=\linewidth,height=4.5cm]{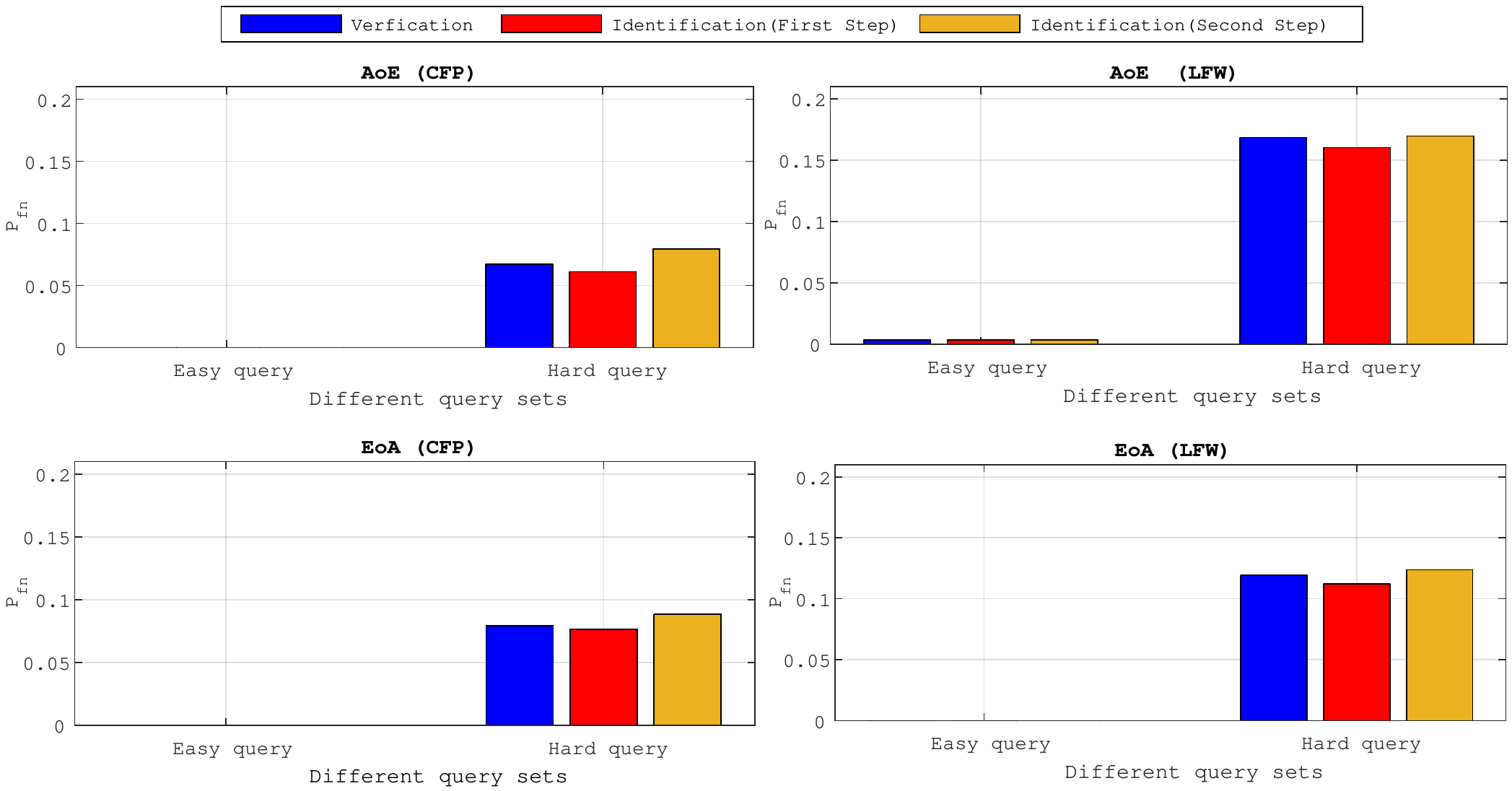}
	\caption{The impact of the similarity of the query with the enrolled template on group verification and identification.}%
	\label{fig:sim}%
\end{figure}

\begin{figure}%
	\centering
	\includegraphics[width=\linewidth]{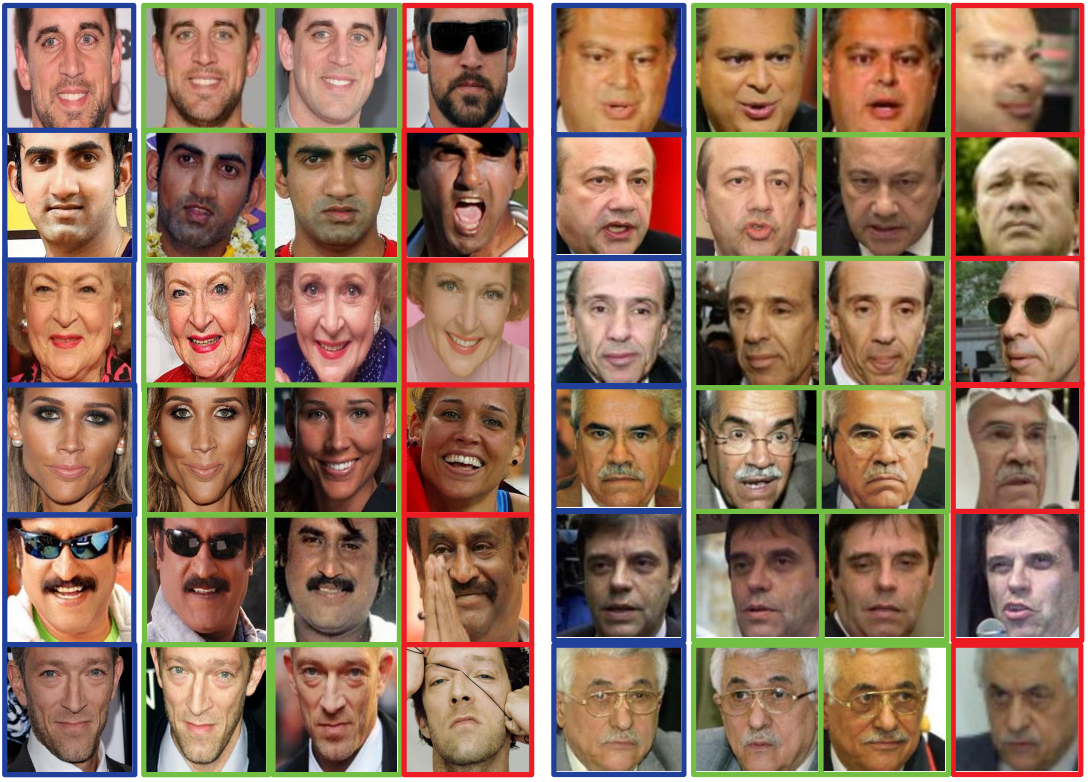}
	\caption{Examples of group identification on CFP(left) and LFW(right). Blue frames indicate enrolled samples, green~/~red frames successful~/~failed queries, respectively.}%
	\label{fig:images}%
\end{figure}

Figure~\ref{fig:sim} gives an additional perspective on the phenomenon highlighted above, that is, the genuine similarity between the query and the enrolled template is a key factor.
Easy queries are very well handled whereas hard queries are more problematic.
Put differently, the proposed method do not severely degrade the recognition power of the descriptors obtained through the VGG16 network. 
Descriptors poorly correlated already at the image level can only cause poor performance once embedded and aggregated.
This is also shown in Fig.~\ref{fig:images} which displays some enrolled and querying faces of the `in the wild' datasets LFW and CFP.  All the failed identification examples show a change of lighting, pose or expression, and~/~or occlusion. Yet, such changes do not automatically give a failure.

\subsection{Exp. \#4: Security and Privacy}
As for the security and privacy, the quantities~\eqref{eq:MSE_P} and~\eqref{eq:MSE_S} were measured as empirical average over the dataset.

\begin{figure}[tb]%
	\centering
	\includegraphics[width=\linewidth,height=4.5cm]{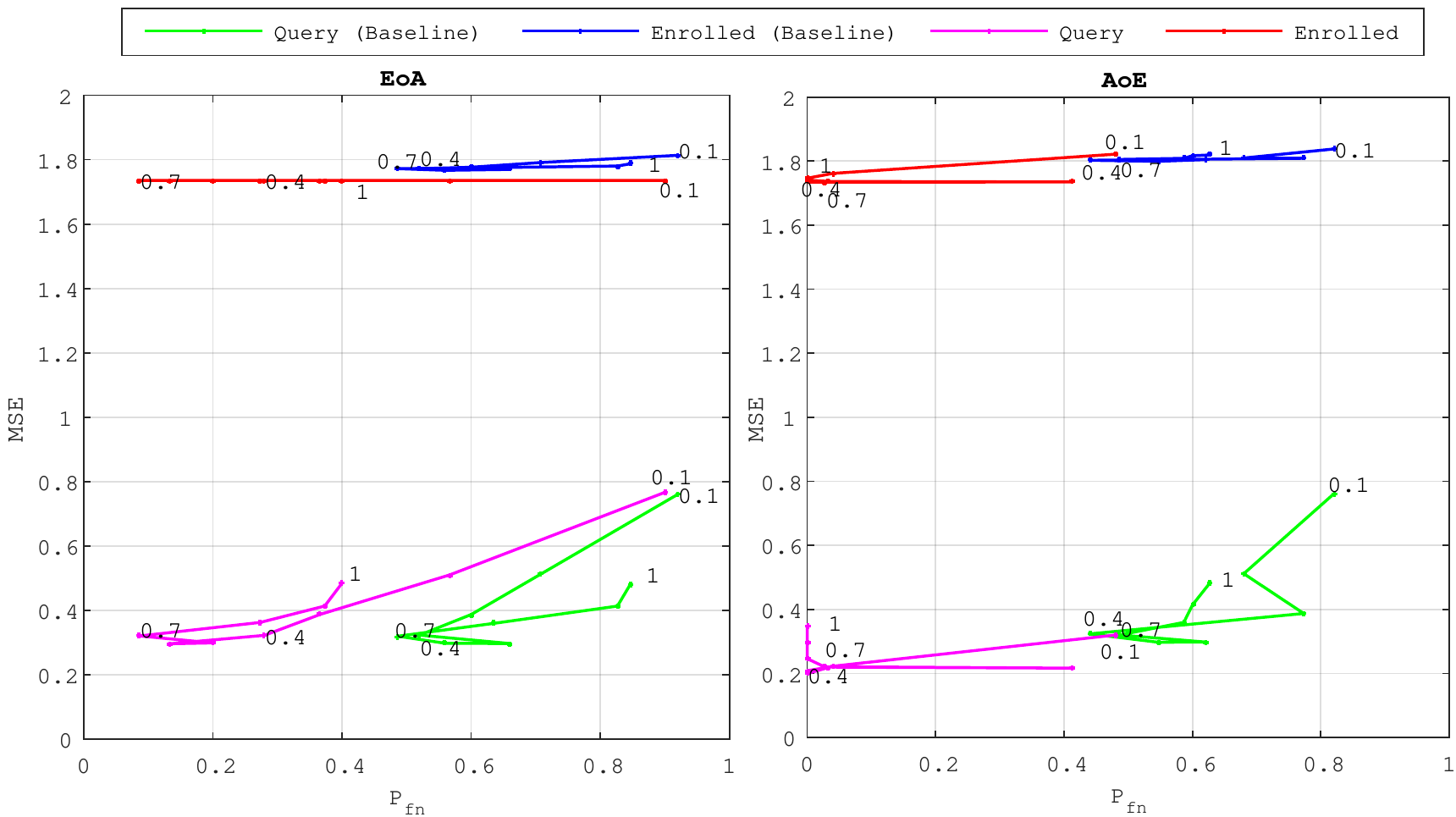}
	\caption{The impact of sparsity factor $S$ on the trade-off between security and performances, on FEI with $m=32$ and $S/d\in(0.1,1)$.}%
	\label{fig:secu}%
\end{figure}

Knowing that the query has unit norm, the reconstruction mechanism yields a unit vector as follows: $\hat{\y} = \W\func{e}(\y)/\|\W\func{e}(\y)\|$.
The quality of the reconstruction mainly depends on the sparsity factor $S$.  When $S$ is small, the template is reconstructed with few columns of $\W$.  When $S$ is big, more columns are used but the amplitude modulating each column is coarsely reconstructed.  There might be two values of $S$, one small, one large, providing the same reconstruction MSE.  However, these two values do not yield the same performances as shown in Fig.~\ref{fig:secu}.  

Reconstructing enrolled templates is even more difficult due to the aggregation (see~\eqref{eq:MSE_S}).  Fig.~\ref{fig:secu} shows that our method has decreased the security a little, but overall the trade-off between security and performances is more interesting especially for AoE.
 

\section{Conclusion}
This paper proposes a framework for group membership verification and identification by jointly learning the embedding and the aggregation.
Yet, this learning was not completely free.
Some guidances were still imposed, especially the prototyping of the embedding based on a sparse ternary quantization. This is mainly for inheriting security and privacy properties of this lossy information processing~\cite{Razeghi2017wifs, Razeghi2018icassp}. It is not clear whether an alternative approach does exist.

Our future work looks at increasing the length of the group representation in order to pack more templates into a group.
Yet, this raises scalability issues appealing for faster but approximative learning.

{\small
\bibliographystyle{ieee}
\bibliography{ms}
}
\end{document}